\documentclass[conference,a4paper]{IEEEtran}
\IEEEoverridecommandlockouts
\pdfoutput=1
\usepackage{amsmath}
\usepackage[noend]{algpseudocode}
\usepackage{graphicx}
\usepackage{url}
\usepackage{pdfpages}
\usepackage{multirow}
\usepackage{verbatim}
\usepackage[vietnamese,english]{babel}
\usepackage[utf8]{inputenc}
\usepackage[linesnumbered,algoruled,boxed,lined]{algorithm2e}
\makeatletter
\def\BState{\State\hskip-\ALG@thistlm}
\makeatother
\DeclareMathOperator*{\argmax}{argmax}

\title{\LARGE \bf
LSTM Easy-first Dependency Parsing with Pre-trained Word Embeddings and Character-level Word Embeddings in Vietnamese
}

\author{\IEEEauthorblockN{Binh Duc Nguyen\IEEEauthorrefmark{1}, Kiet Van Nguyen\IEEEauthorrefmark{2} and Ngan Luu-Thuy Nguyen\IEEEauthorrefmark{3}}\IEEEauthorblockA{University of Information Technology \\Vietnam National University – Ho Chi Minh City, Vietnam \\ Email: \IEEEauthorrefmark{1}14520071@gm.uit.edu.vn, \IEEEauthorrefmark{2}kietnv@uit.edu.vn, \IEEEauthorrefmark{3}ngannlt@uit.edu.vn}}

\begin{document}

\maketitle
\thispagestyle{empty}
\pagestyle{empty}

\begin{abstract}
In Vietnamese dependency parsing, several methods have been proposed. Dependency parser which uses deep neural network model has been reported that achieved state-of-the-art results. In this paper, we proposed a new method which applies LSTM easy-first dependency parsing with pre-trained word embeddings and character-level word embeddings. Our method achieves an accuracy of 80.91\% of unlabeled attachment score and 72.98\% of labeled attachment score on the Vietnamese Dependency Treebank (VnDT).


\end{abstract}

\section{Introduction}
Over the last decade, there has been considerable interest in dependency parsing which generates grammatical relations between two words in the sentence. For instance, in the CoNLL 2007 shared task \cite{conll}, each participating team tested their system in thirteen different languages. In 2014, the SPMRL shared task \cite{spmrl} was held to evaluate dependency parsing on nine morphologically rich languages.

Since CoNLL-X \cite{conllx}, there are two dominant dependency parsing models which are transition-based parsing and graph-based parsing. These names are first mentioned in \cite{data-driven}. Both two models are data-driven parsers which are learned from an annotated corpus and have shorter development time than rule-based system \cite{data-driven}.

The last few years has witnessed a rapid development of neural-network methods. Many works showed that neural-network methods can achieve state-of-the-art results on different tasks of Natural Language Processing such as Named Entity Recognition (NER) \cite{ner1}, Machine Translation \cite{ml1}, \cite{ml2}. In dependency parsing,  there are studies using deep neural-networks to encode features without hand-crafted definition \cite{char:dyerb}, \cite{stacklstm}, \cite{kis:a} and \cite{kis:b}.

In Vietnamese, there are several researchers working on dependency parsing. Nguyen et al. \cite{treebank1} and Nguyen et al. \cite{treebank2} automatically convert a constituent treebank to a dependency treebank. Nguyen and Nguyen \cite{kiet:b} used supertags features and Vu-Manh et al. \cite{wordemb} used word embeddings on the transition-based parser. Nguyen et al \cite{bistvn} used BiLSTM encoder to generate word representation vectors and obtained a state-of-the-art result on Vietnamese Dependency Treebank (VnDT) \cite{treebank2}.

We found that \cite{bistvn} did not incorporate pre-trained word embeddings in the vector representation of words. Although this feature can improve the accuracy of dependency parsing because of its rich context information \cite{kis:a} \cite{kis:b} and  \cite{wordemb}.

Beside the pre-trained word embeddings, character-level word embeddings has been found useful for NER \cite{ner1} or dependency parsing \cite{char}. It represents words by combining their character embeddings so it does not depend on the word-based lookup dictionary. However, no previous study has investigated the impact of character-level word embeddings on dependency parsing for Vietnamese.

The parser we used is a variant of transition-based parser called tree LSTM easy-first \cite{kis:a} which is fast speed and similar to human annotation \cite{goldberg}. The tree LSTM easy-first is reported as the state-of-the-art parser for Chinese \cite{kis:a}. Because Chinese syntax structures is very similar to Vietnamese, we believe that this parser can give a good accuracy for Vietnamese dependency parsing.

In this paper, we focused on applying two word representation features on Vietnamese dependency parsing, which are pre-trained word embeddings and character-level word embeddings. Although Vu-Manh et al. \cite{wordemb} have already used pre-trained word embeddings for Vietnamese dependency parsing. Their paper used MaltParser \cite{malt} which is incomparison with other neural network parsers like Tree LSTM easy-first parser. Our paper is the first study which applied character-level word embedding to dependency parsing on VnDT as well as the first time easy-first parser was utilized with character-level word embeddings in either Vietnamese or English. Our method makes an improvement in parsing accuracy (0.79\% UAS and 1.51\% LAS) and achieves state-of-the-art performance on Vietnamese dependency parsing (80.91\% UAS and 72.98\% LAS).

The remainder of the paper is organized as follows: Section 2 describes easy-first parsing algorithm used in this paper. Word representation features are represented in Section 3. We evaluate the results of the methods in Section 4 and draw the conclusion and future work in Section 5.
\section{Easy-first Parsing}
Easy-first parsing is a type of syntactic parsing algorithm which is a variant of transition-based parsing. This method is first proposed by Goldberg and Elhadad \cite{goldberg}. Several studies were proposed for improving the accuracy of this method such as Ma et al \cite{jma} using beam search strategy to effectively explore the search space of parsing process. Another method was proposed by Kiperwasser \& Goldberg \cite{kis:a} which uses deep neural network to learn model's parameters, entire dependency tree was encoded by LSTMs and was applied as deep features for effective learning.
\subsection{The parsing process}
\begin{figure}[h]
\includegraphics[page=1, width=80mm]{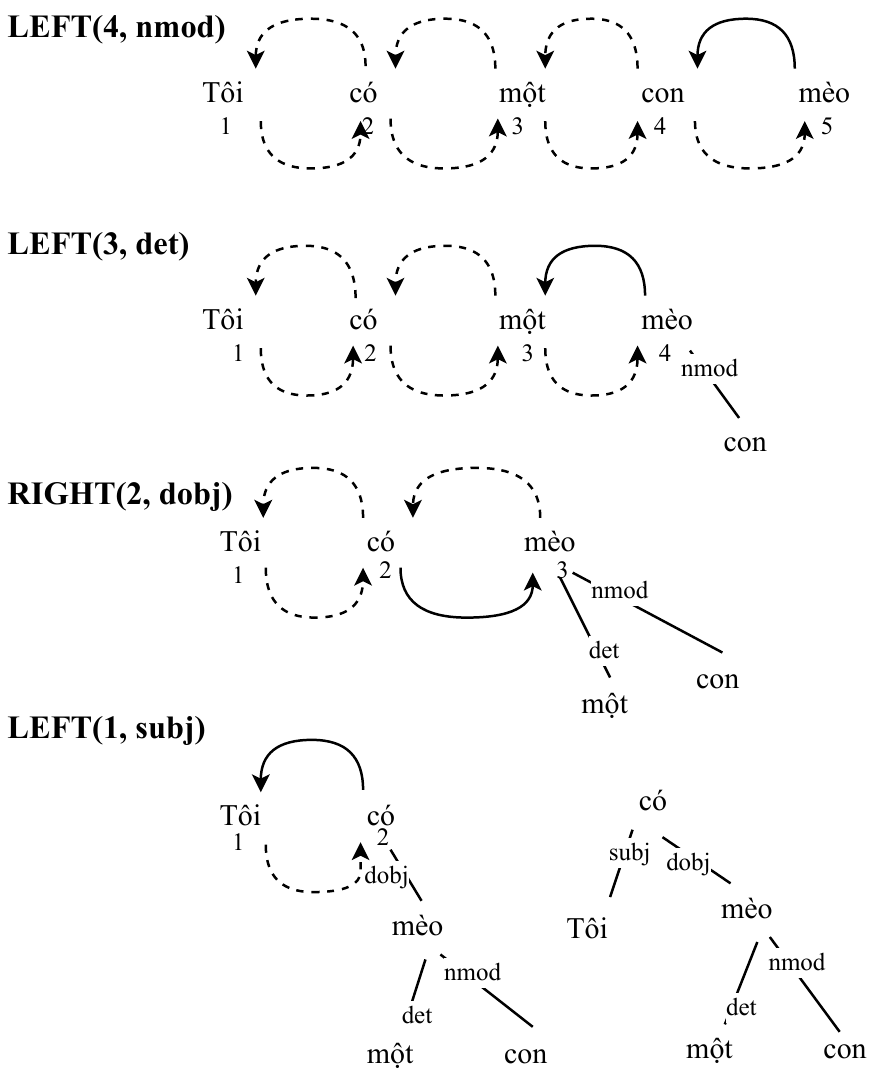}
\centering

\caption{\protect\begin{otherlanguage}{vietnamese}
		The sentence "Tôi có một con mèo" is parsed using easy-first algorithm.
	\protect\end{otherlanguage}}

\label{fig1}
\end{figure}

\begin{otherlanguage}{vietnamese}
In the parsing process, there are a list of partial structures called \textit{pending} which is the main data-structure of the parser, the parser will stop if the \textit{pending} have only one element in it. A partial structure can be a token or a dependency structure which is built from the previous parse steps. At each step of the parsing process, the algorithm generates all possible actions. These actions are ranked by a scoring function. The highest scoring action is chosen to applied to the \textit{pending}. There are two types of actions which are \textbf{LEFT} and \textbf{RIGHT}. Let  $p_1, p_2, p_3 ..., p_n$ be the elements of \textit{pending}. The action \textbf{LEFT(\textit{i,r})} adds the dependency edge $(p_{i+1},  p_i)$ with the relation \textit{r} and remove $p_i$ from the \textit{pending}, while the action \textbf{RIGHT(\textit{i,r})} adds the  dependency edge $(p_i, p_{i+1})$ with the relation \textit{r} and remove $p_{i+1}$ from the \textit{pending}. Figure \ref{fig1} demonstrates how to parse the sentence \textit{\protect\begin{otherlanguage}{vietnamese}"Tôi có một con mèo"\protect\end{otherlanguage}} using easy-first algorithm. Let Arcs be the list of dependency edges in parsing process. At step 1, action \textbf{LEFT(\textit{4, nmod})} is chosen, edge \textit{(mèo, con, nmod)} is added to Arcs. At step 2, action \textbf{LEFT(\textit{3,det})} is ranked highest so the edge \textit{(mèo, một, det)} is added to Arcs. At the third step, the chosen action is \textbf{RIGHT(\textit{2, dobj})} and the edge \textit{(có, mèo, dobj)} is added to Arcs. At the final step, action \textbf{LEFT(\textit{1,nsubj})} was chosen and produce the output dependency structure.
\end{otherlanguage}

At one step of the parsing process, for a \textit{pending} with $n$ elements, there are $n-1$ attachment points where actions can be applied to. For each attachment point, there will be $2*R$ actions where $R$ is the number of total dependency relations. Therefore, there are $2R(n-1)$ actions can be chosen at each step so there are more than one valid action sequences leads to the dependency structure of the sentence. That gives more examples to the learning algorithm.

\subsection{The LSTMs Easy-first algorithm}
\begin{figure}[h]
\includegraphics[page=1, width=80mm]{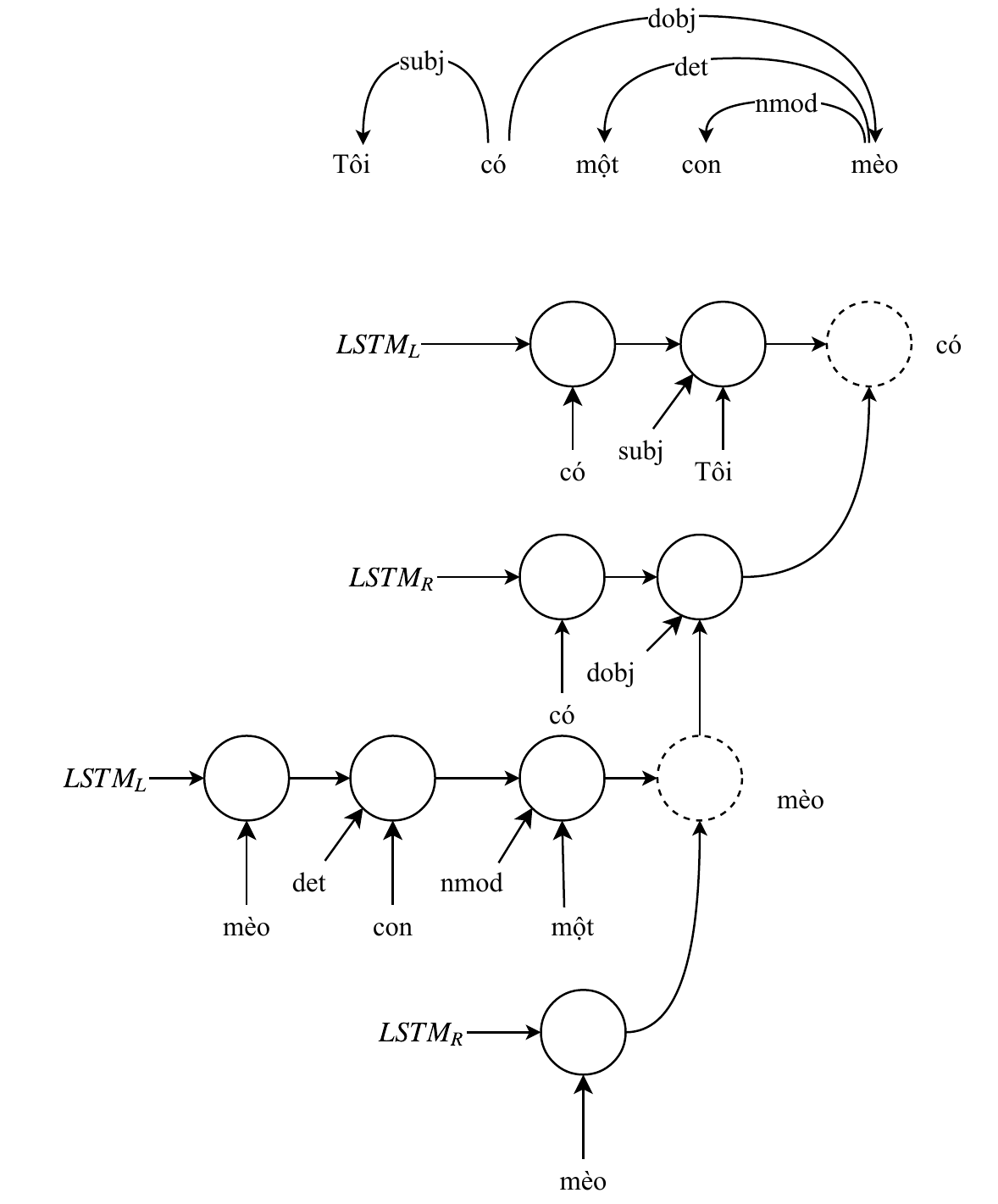}
\centering
\caption{
	\protect\begin{otherlanguage}{vietnamese}
		Networks of dependency structure of sentence "Tôi có một con mèo". The solid circle is is a representation of  the hidden state of LSTM networks, the dotted circle is a representation of the function of concatenating two LSTM networks. The inputs of LSTM hidden state include previously hidden state, vector representation of the child node and a vector represented for dependency relation between child node and head node.
	\protect\end{otherlanguage}}
\label{fig2}
\end{figure}

In this paper, we used an extended version of the easy-first algorithm \cite{kis:a}, which employs LSTMs to represent a dependency structure as a vector and used multilayer perceptron (MLP) to score the parse actions. The dependency structure can be described as follows \cite{kis:a}:
\begin{itemize}
\item[]
\includegraphics[page=1, width=80mm]{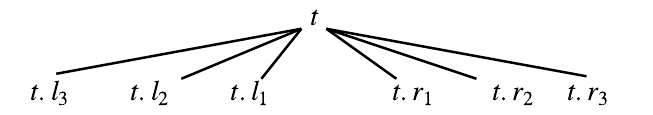}

\item The depenency structure is a tree with root node $t$ and $t$ is associated with the head word $w_t$.

\item For every child of $t$, if the \textit{head word} of this child is on the left of $w_t$ according to their positions in the sentence, this child will be the left child, otherwise, this child will be the right child. The nearest child of the head node is indexed as 1 while the left most or right most child has the largest index.

\item Let $enc(t)$ be the vector representing dependency structure $t$. All of left children $t.l_1 ... t.l_{k_l}$ are fed to an LSTM called $LSTM_L$, all of right children $t.r_1 ... t.r_{k_r}$ are fed to an LSTM called $LSTM_R$. The first input of $LSTM$ is the vector representation of the head word $v_i(t)$, the last input is the vector representation of left-most child or right-most child. The output of $LSTM_L$ and $LSTM_R$ are concatenated $e_l(t) \circ e_r(t)$. The dimension of the concatenated vector is reduced using linear transformation, followed by a non-linear activate function. The result vector represents dependency structure.
\begin{equation}
enc(t) = g(W^e . (e_l (t) \circ e_r (t) \circ l(t))+b^e)
\end{equation}
\begin{equation}
e_l(t) = LSTM_L (v_i(t), enc(t.l1)... enc(t.l_{k_l}))
\end{equation}
\begin{equation}
e_r(t) = LSTM_R (v_i(t), enc(t.r1)... enc(t.r_{k_r}))
\end{equation}

\item The process runs recursively and stop at leaf nodes where $v_i(leaf)$ is the vector representation of word $i$ in sentence. Which will be described in the Section 3. Figure \ref{fig2} shows the network of dependency structure of the sentence \begin{otherlanguage}{vietnamese}"Tôi có một con mèo"\end{otherlanguage}.
\begin{equation}
enc(leaf) = g(W^e . (e_l (leaf) \circ e_r (leaf))+b^e)
\end{equation}

\begin{equation}
e_l(leaf) = LSTM_L (v_i(leaf))
\end{equation}

\begin{equation}
e_r(leaf) = LSTM_R (v_i(leaf))
\end{equation}

\end{itemize}

\begin{algorithm}
\caption{LSTM Easy-first dependency parsing algorithm}\label{alg2}
\SetAlgoLined
\SetKwInOut{Input}{Input}\SetKwInOut{Output}{Output}
\Input{a sentence= $w_1...w_n$, parameter $w$}
\Output{sentence's dependency arcs}
$Arcs \leftarrow \{\}$ \;
\For{i in \{0..len(sentence)-1\}}{
	$pending[i].w = w_i$\;
	$pending[i].LSTM_L.init().append(w_i)$\;
	$pending[i].LSTM_R.init().append(w_i)$\;
}
\While{$len(pending) > 1$}{
	$actions \leftarrow \{\}$\;
	\For{i in $0 ... len(pending) - 1$}{
		\For{r in $Rels$}{
			\For{a in $[LEFT, RIGHT]$}{
				$actions.append(i, a, r)$\;
			}		
		}	
	}\
	(i, a, r) = $\argmax_{act \in actions} Score(act, w)$\;
	\uIf{a == LEFT}{
		$Arcs.append(p_{i+1}, p_i, r)$\;
		$pending[i+1].LSTM_L.append(p_i, r)$\;
		$pending.remove(p_i)$ \;
	}\uElseIf{a == RIGHT}{
		$Arcs.append(p_i, p_{i+1}, r)$\;
		$pending[i].LSTM_R.append(p_{i+1}, r)$\;
		$pending.remove(p_{i+1})$ \;
	}
}
\Return{Arcs}
\end{algorithm}

Pseudocode for Tree LSTM Easy-first parsing is described in Algorithm \ref{alg2} \cite{kis:a}. The parsing algorithm is similar to the original \cite{goldberg} but with some differences:
\begin{itemize}
	\item When the action was applied to $pending$, the vector of modifier will be appended to the LSTMs of the head ($LSTM_L$ or $LSTM_R$).
	
	\item There are two scoring functions which are modeled as multi-layer perceptrons (MLP) - $Score_U$ and $Score_R$. $Score_U$ scores an action based on its head and modifier while $Score_R$ scores an action based on head, modifier and the relation of this action. Scoring functions are using information of partial structure at the attachment point as well as its neighbors. The score of an action $a$ with relation $r$ in $Score_U$ or $Score_R$ is an element of the output layer of MLP. Size of the output layer of $MLP_U$ is 2 while $MLP_R$ is $2R$ where $R$ is the total number of dependency relations:
\begin{equation}
Score_U(i, a) = MLP_U(x_i)[a]
\end{equation}

\begin{equation}
Score_L(i, a, r) = MLP_L(x_i)[r,a]
\end{equation}

\begin{equation}
x_i = p_{i-2} \circ ... \circ p_{i+3}
\end{equation}

\item The score of an action is the sum of $Score_U$ result and $Score_R$ result:
\begin{equation}
Score(i, a, r) = Score_U(i, a) + Score_L(i, a, r)
\end{equation}

\end{itemize}

\subsection{The training process}
For parameters update, a hinge loss function with margin 1 is used \cite{kis:a}:

\begin{equation}
\begin{split}
max\{0,1 - max_{i, a, r \in G}Score(i, a, r) + \\
max_{i, a, r \in A  \setminus  G}Score(i, a,  r)\}
\end{split}
\end{equation}

Pseudocode of training phrase is described in Algorithm \ref{alg3} \cite{kis:a}. Where $A$ is set of valid actions and $G$ is set of all possible actions at the current step. For validation checking, the parser uses \textit{Oracle} which is a set of defined rules based on gold dependency structure of a sentence. The \textit{Oracle} used in Tree LSTM Easy-first is \textit{dynamic oracle} \cite{dynamic}. If an invalid action was chosen, in the next step, the parser will treat an action which leads to best acceptable dependency structure as the valid action. The rules used in Tree LSTM Easy-first parser are defined as follows:
\begin{itemize}
	\item \textit{Head, modifier and relation} of action must be in the gold dependency structure of the sentence.
	\item \textit{Modifer} of the action must be complete which means that all children of the modifier have been explored at the previous step.
	\item If the gold head of modifier is removed from pending, the action then is valid if the  dependency relation of modifier is the same with the relation in the gold dependency structure despite that the head is different.
\end{itemize}

\begin{algorithm}[h]
\caption{LSTM Easy-first training algorithm for dependency parsing}\label{alg3}
\SetAlgoLined
\SetKwInOut{Input}{Input}\SetKwInOut{Output}{Output}
\Input{a sentence= $w_1...w_n$, parameter $w$}
\Output{sentence's dependency arcs}
\For {i in 1...n}{
$errors \leftarrow []$ \;
\For {sentence in corpus}
{
\For{i in 0..len(sentence)-1}{
	$pending[i].w = w_i$\;
	$pending[i].LSTM_L.init().append(w_i)$\;
	$pending[i].LSTM_R.init().append(w_i)$\;
}
\While{$len(pending) > 1$}{
	$actions \leftarrow \{\}$\;
	\For{i in \{$0 ... len(pending) - 1$\}}{
		\For{r in $Rels$}{
			\For{a in $[LEFT, RIGHT]$}{
				\uIf{is\_valid(i, a, r)} {
					$G.append(i, a, r)$\;				
				}\uElse{
					$A.append(i, a, r)$\;}
			}		
		}	
	}\
	$i, a, r, s$ = $\argmax_{act \in A} Score(act, w)$\;
	$i', a', r', s'$ = $\argmax_{act \in G} Score(act, w)$\;
	\uIf {$s > 1 + s'$}{
		$i_{best}, a_{best}, r_{best} = i, a, r$\;
	}
	\uElse{
		$i_{best}, a_{best}, r_{best} = i', a', r'$\;
		$errors.append(1 - s + s')$\;	
	}
	$apply\_action(i_{best}, a_{best}, r_{best})$\;
	\If{$len(errors) > 50$}{
		update\;
		$errors \leftarrow []$\;	
	}	
	
	}}
\If{$len(errors) > 0$}{
		update\;
}	
}
\end{algorithm}
When the loss of parse step is greater than zero, this step is count as one error. If the total errors are greater than 50, all parameters are updated using Adam Optimizer.

\section{Words representation}
\subsection{Representing words using word form and POS tag}
\begin{figure}[h]
\includegraphics[page=1, width=80mm]{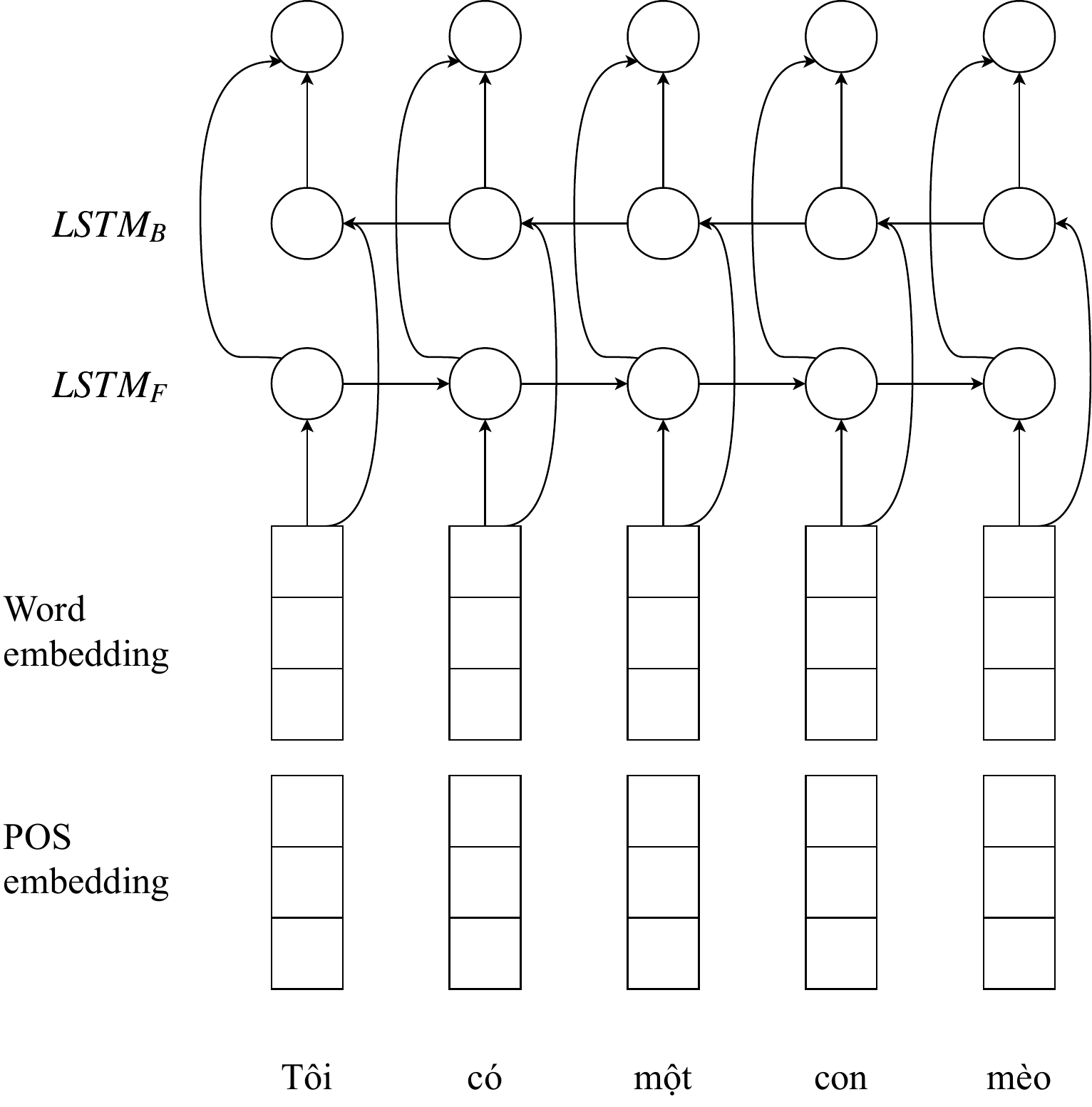}
\centering
\caption{Words's vector representation using BiLSTMs}
\label{fig3}
\end{figure}

The baseline approach of representing words as vectors is used information of word form and POS tag which are embeddings and jointly trained with the networks. Word-form and POS-tag vectors are concatenated and transformed to $v'_i$ via a linear transformation followed by a non-linear activation function. Kiperwasser and Goldberg \cite{kis:a} used Bidirectional LSTMs to incorporate the context information of words in a sentence. For the word $i^{th}$ in sentence, network $LSTM_F$ runs from the begining of the sentence to word $i^{th}$ while $LSTM_B$ runs in reverse order. Outputs of two networks are concatenated to produce vector $v_i$ representing word $i$. The figure \ref{fig3} illustrates how a word is represented using BiLSTMs.
\begin{equation}
v'_i = g(W^v .(w_i \circ p_i) + b^v)
\end{equation}
\begin{equation}
f_i = LSTM_F(v'_1, v'_2,...v'_i)
\end{equation}
\begin{equation}
b_i = LSTM_B(v'_n, v'_{n-1},..., v'_i)
\end{equation}
\begin{equation}
v_i = (f_i \circ b_i)
\end{equation}
\subsection{Pretrained word embeddings}
A common way to incorporate context information of words is using word representations learned from unannotated corpora. In pre-trained word embeddings, each word in vocabulary is associated with a high dimensional real-valued vector which is learned from neural network model. Vocabulary can be considered as points in vector space and words which are similar in meaning are closer in vector space. The representation can capture syntactic and semantic relationships between words. To investigate the effect of pre-trained word embedding, we used two existing pre-trained word embedding models on Vietnamese (skip-gram \cite{skip} and subwords \cite{sub}) and used them as additional information along with word form embedding and POS tag embedding:

\begin{equation}
v'_i = g(W^v .(w_i \circ p_i \circ extn_i) + b^v)
\end{equation}
\subsection{Character-level word embedding}
The main contribution of this study is using character-level word embeddings as additional information of word representation in the easy-first algorithm and applied it on VnDT. Unlike pre-trained word embeddings, character-level word embeddings can deal with out-of-vocabulary (OOV) problem because unknown words can be generated using their component characters. As characters are shared across words, character embedding combination still presents semantic information of words like pre-trained words embeddings.

\begin{figure}[h]
\includegraphics[page=1, height=60mm]{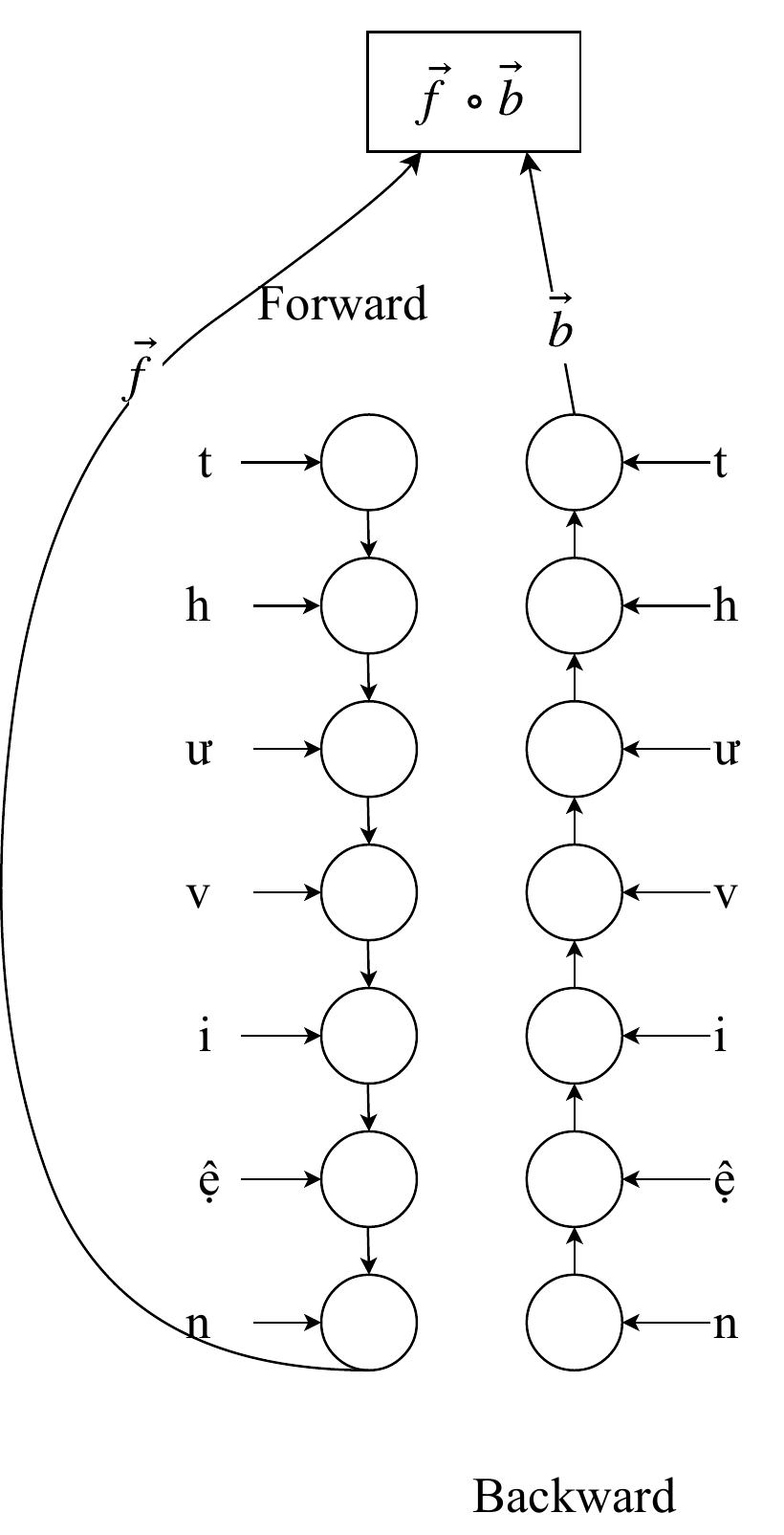}
\centering
\caption{
	\protect\begin{otherlanguage}{vietnamese}
		The word "thư viện" is represented by character embedding with BiLSTM
	\protect\end{otherlanguage}}
\label{char}
\end{figure}

We used the same network structure as in \cite{char}. Character embeddings of words is fed to the bidirectional LSTMs which include two LSTMs called \textbf{Forward} and \textbf{Backward}. The input of the forward network is the character embeddings of the characters from the beginning to the end of the word while inputs of the backward network is characters of the word in the reverse order. The output of the two networks is concatenated to one vector which represents information of words. Character embeddings are learned jointly with the training process. The figure \ref{char} shows an example of using character embeddings for representing a word.

The hyper-parameters of the networks used for training character embeddings are detailed in Table I.

\begin{table}[h]
	\label{charparams}
	\centering
	\caption{hyper-parameters used for character embedding networks}
	\begin{tabular}{|l|l|}
		\hline
		Char embedding dimension & 100       \\ \hline
		BI-LSTM Layers           & 2         \\ \hline
		BI-LSTM Dimensions       & 100 + 100 \\ \hline
	\end{tabular}
\end{table}

\section{Results and Discussion}
In our experiments, we use the VnDT \cite{treebank2} which  has 10,200 dependency structures. For comparison purposes, we split the data in the same way as \cite{bistvn}, last 1020 sentences for testing (POS and automatic POS tagging) and the rest for training. The POS tagging tool we used is VnTagger \footnote{https://github.com/scorpion1206/VnTagger}, its accuracy is 94.4\% on our test set. The performance is measured with unlabeled attachment score (\textbf{UAS}) and  labeled attachment score (\textbf{LAS}).

We used two pre-trained word embeddings datasets which were trained with two different models on Vietnamese: skip-gram\footnote{https://github.com/pth1993/NNVLP} and subwords\footnote{https://github.com/facebookresearch/fastText}. Information on these two datasets is described in table \ref{pretrained}. Coverage column shows the percentage of words in the VnDT treebank which appears in pre-trained word embeddings.

Because of the difference between word tokenization methods, many words in VnDT do not exist in pre-trained word embeddings, this is the reason for the low percentage of vocabulary coverage of both datasets. If a word cannot be found in pre-trained word embeddings, this word will be treated as an unknown word and represented by the \textit{unknown} vector. This unknown vector do not carry much useful context information of words that leads to poor performance in parsing.\\
\begin{table}[h]
\centering
\caption{Information of two Pre-trained word embeddings}
\label{pretrained}
\begin{tabular}{|c|c|c|c|}
\hline
\textbf{Model} & \textbf{Dim} & \textbf{Vocabulary} & \textbf{Coverage} \\ \hline
Subword        & 300          & 200,000              & 20\%              \\ \hline
Skip-gram      & 300          & 100,000              & 67\%              \\ \hline
\end{tabular}
\end{table}

For fully exploration of pre-trained word embeddings and character-level word embeddings, we design our experiments as follows:
\begin{itemize}
\item Word embeddings + POS tag embeddings.
\item Word embeddings + POS tag embeddings + Character embeddings.
\item Word embeddings + POS tag embeddings + Pre-trained word embeddings.
\item Word embeddings + POS tag embeddings + Character embeddings + Pre-trained word embeddings.
\end{itemize}

We also compare the result of the easy-first algorithm with the current state-of-the-art parsers proposed by Kiperwasser and  Goldberg \cite{kis:b} which uses BiLSTM encoder to represent words in the sentence, the word vector is used as features in transition-based parser (BistT) and graph-based parser (BistG) \cite{kis:a} \footnote{https://github.com/elikip/bist-parser}. The implementation of Tree LSTM easy-first is provided at \url{https://github.com/elikip/htparser}. The results are demonstrated in Table \ref{result}. In the table, the results of BistT and BistG are taken from [19] because this is reported as the state-of-the-art parser on VnDT.
\begin{table}[h]
\centering
\caption{Vietnamese dependency parsing results on different parsers}
\label{result}
\begin{tabular}{|l|l|l|l|l|}
\hline
\multicolumn{1}{|c|}{\multirow{2}{*}{Model}} & \multicolumn{2}{c|}{Gold POS Tag}                           & \multicolumn{2}{c|}{Auto POS Tag}                           \\ \cline{2-5} 
\multicolumn{1}{|c|}{}                       & \multicolumn{1}{c|}{UAS\%} & \multicolumn{1}{c|}{LAS\%} & \multicolumn{1}{c|}{UAS\%} & \multicolumn{1}{c|}{LAS\%} \\ \hline
BistT                           & 79.33                    & 72.53                    & 76.56                    & 68.22               
\\ \hline
BistG                          & 79.39                    & 73.17                    & 76.28                    & 68.40                    \\ 
\hline
Easy-first (Word + POS)                      & 79.29                    & 71.44                    & 77.51                    & 68.27                    \\
+ Char                                       &         80.58                 &                         72.95  &                 78.01         &           68.82               \\ 
+ Skip-gram                                    & 79.69                    & 72.23                    & 76.91                    & 67.77                    \\ 
\quad + Char                                       &                         \textbf{80.91}   &             \textbf{72.98}           &                         \textbf{78.26} &                \textbf{69.04}          \\ 
+ Sub-word                                  & 79.50                    & 71.94                    & 77.16                    & 67.96                    \\
\quad + Char                                       &                         80.56 & 72.37                          &         78.13                 &       68.56                   \\
\hline
\end{tabular}
\end{table}

Using character-level word embeddings as additional information improve the accuracy of easy-first parser (0.79\% UAS and 1.51\% LAS). Replacing word embedding with character-level word embedding shows slight increase in the performance of the parser (0.39\% UAS and 0.09\% LAS). These results show that using character-level word embeddings can improve the accuracy of parsing.

Using both skip-gram and subword pre-trained word embeddings increase the accuracy of the parser. Despite that the difference of vocabulary coverage between skip-gram and subword is high (40\%), the accuracy of the parser using skip-gram is only higher than subword  0.16\%. We found that in our test data, the different of token coverage (percentage of token in test data which appears in pre-trained word embeddings) between two datasets is only 5\% (75.39\% for skip-gram and 70.95 \% for subword).

Among all features we applied to Tree LSTM easy-first parser, the combination of \textit{word embedding + POS-tag embedding + character-level word embedding + pre-trained word embedding} gives the best parse result (80.91\% UAS and 72.98\% LAS). Our model has outperformed the BistG parser and obtained state-of-the-art performance on the VnDT.

\section{Conclusion}
We have demonstrated the effectiveness of pre-train word embedding and character-level word embedding as features for Vietnamese easy-first dependency parsing. 

In future work, we would like to train character-level word embedding on larger corpora and use it as pre-trained features like pre-trained word embeddings.

\section{Acknowledgement}
This research is funded by University of Information Technology - Vietnam National University Ho Chi Minh City under grant number D1-2017-13.


\begin{thebibliography}{99}
\bibitem{conll} Nivre, Joakim, et al. "The CoNLL 2007 shared task on dependency parsing." Proceedings of the 2007 Joint Conference on Empirical Methods in Natural Language Processing and Computational Natural Language Learning (EMNLP-CoNLL). 2007.
\bibitem{spmrl} Seddah, Djamé, Sandra Kübler, and Reut Tsarfaty. "Introducing the spmrl 2014 shared task on parsing morphologically-rich languages." Proceedings of the First Joint Workshop on Statistical Parsing of Morphologically Rich Languages and Syntactic Analysis of Non-Canonical Languages. 2014.
\bibitem{conllx} Buchholz, Sabine, and Erwin Marsi. "CoNLL-X shared task on multilingual dependency parsing." Proceedings of the Tenth Conference on Computational Natural Language Learning. Association for Computational Linguistics, 2006.
\bibitem{data-driven} McDonald, Ryan, and Joakim Nivre. "Characterizing the errors of data-driven dependency parsing models." Proceedings of the 2007 Joint Conference on Empirical Methods in Natural Language Processing and Computational Natural Language Learning (EMNLP-CoNLL). 2007.
\bibitem{ner1} Chiu, Jason PC, and Eric Nichols. "Named entity recognition with bidirectional LSTM-CNNs." arXiv preprint arXiv:1511.08308 (2015).
\bibitem{ml1} Bahdanau, Dzmitry, Kyunghyun Cho, and Yoshua Bengio. "Neural machine translation by jointly learning to align and translate." arXiv preprint arXiv:1409.0473 (2014).
\bibitem{malt}Nivre, Joakim, Johan Hall, and Jens Nilsson. "Maltparser: A data-driven parser-generator for dependency parsing." Proceedings of LREC. Vol. 6. 2006.
\bibitem{ml2} Sutskever, Ilya, Oriol Vinyals, and Quoc V. Le. "Sequence to sequence learning with neural networks." Advances in neural information processing systems. 2014.
\bibitem{char:dyerb} Dyer, C., Ballesteros, M., Ling, W., Matthews, A., \& Smith, N. A. (2015). Transition-based dependency parsing with stack long short-term memory. arXiv preprint arXiv:1505.08075.
\bibitem{stacklstm} Ballesteros, M., Dyer, C., Goldberg, Y., \& Smith, N. A. (2017). Greedy transition-based dependency parsing with stack lstms. Computational Linguistics, 43(2), 311-347.
\bibitem{kis:a} Kiperwasser, Eliyahu, and Yoav Goldberg. "Easy-first dependency
parsing with hierarchical tree LSTMs." arXiv preprint arXiv:1603.00375 (2016).
\bibitem{kis:b} Kiperwasser, Eliyahu, and Yoav Goldberg. "Simple and accurate dependency parsing using bidirectional LSTM feature representations." arXiv preprint arXiv:1603.04351 (2016).
\bibitem{char} Ballesteros, Miguel, Chris Dyer, and Noah A. Smith. "Improved transition-based parsing by modeling characters instead of words with lstms." arXiv preprint arXiv:1508.00657 (2015).
\bibitem{treebank1} Thi, Luong Nguyen, et al. "Building a treebank for Vietnamese dependency parsing." Computing and Communication Technologies, Research, Innovation, and Vision for the Future (RIVF), 2013 IEEE RIVF International Conference on. IEEE, 2013.
\bibitem{treebank2} Nguyen, D. Q., Nguyen, D. Q., Pham, S. B., Nguyen, P. T., \& Le Nguyen, M. (2014, June). From treebank conversion to automatic dependency parsing for Vietnamese. In International Conference on Applications of Natural Language to Data Bases/Information Systems (pp. 196-207). Springer, Cham.
\bibitem{kiet:b} Nguyen, Kiet V., and Ngan Luu-Thuy Nguyen. "Vietnamese transition-based dependency parsing with supertag features." Knowledge and Systems Engineering (KSE), 2016 Eighth International Conference on. IEEE, 2016. 175-180). IEEE.
\bibitem{wordemb} Vu-Manh, C., Luong, A. T., \& Le-Hong, P. (2015, December). Improving Vietnamese dependency parsing using distributed word representations. In Proceedings of the Sixth International Symposium on Information and Communication Technology (pp. 54-60). ACM.
\bibitem{bistvn} Nguyen, D. Q., Dras, M., \& Johnson, M. (2016). An empirical study for Vietnamese dependency parsing. Proceedings of the Australasian Language Technology Association Workshop 2016, 143–149. Retrieved from http://www.aclweb.org/anthology/U16-1017
\bibitem{goldberg} Goldberg, Yoav, and Michael Elhadad. "An efficient algorithm for easy-first non-directional dependency parsing." Human Language Technologies: The 2010 Annual Conference of the North American Chapter of the Association for Computational Linguistics. Association for Computational Linguistics, 2010.
\bibitem{jma} Ma, Ji, et al. "Easy-first pos tagging and dependency parsing with beam search." Proceedings of the 51st Annual Meeting of the Association for Computational Linguistics (Volume 2: Short Papers). Vol. 2. 2013.
\bibitem{char:dyera} Ballesteros, M., Dyer, C., \& Smith, N. A. (2015). Improved transition-based parsing by modeling characters instead of words with lstms. arXiv preprint arXiv:1508.00657.
\bibitem{dynamic} Goldberg, Y., \& Nivre, J. (2012). A dynamic oracle for arc-eager dependency parsing. Proceedings of COLING 2012, 959-976.
\bibitem{skip} Mikolov, Tomas, et al. "Distributed representations of words and phrases and their compositionality." Advances in neural information processing systems. 2013.
\bibitem{sub} Bojanowski, Piotr, et al. "Enriching word vectors with subword information." arXiv preprint arXiv:1607.04606 (2016).

\end{thebibliography}
\end{document}